\documentclass{article}
\usepackage[utf8]{inputenc}
 \usepackage[T1]{fontenc}
\usepackage{amsmath}
\usepackage{pifont}
\usepackage{bm}
 \usepackage{amssymb}
 \usepackage{amsthm}
\usepackage{parskip}
\usepackage[ruled, noend, linesnumbered]{algorithm2e}
\usepackage[colorlinks = true, pdfstartview = FitV, linkcolor = blue, citecolor = blue, urlcolor = blue, bookmarks = false]{hyperref}
\usepackage[authoryear, round]{natbib}
\usepackage{color,xcolor}
\usepackage{setspace} 
\usepackage{graphicx}
\usepackage{subcaption}
\usepackage{setspace}
\usepackage{xcolor}
\usepackage{caption}
\usepackage{tabularray}
\UseTblrLibrary{booktabs}

\usepackage[margin=1.315
in]{geometry}
\newcommand{\m}[1]{\mathrm{#1} }

\renewcommand{\v}[1]{\boldsymbol{#1}}
\newcommand{\bb}[1]{\mathbb{#1}}

\newcommand{\ignore}[1]{}

\newsavebox\IBoxA \newsavebox\IBoxB \newlength\IHeight
\newcommand\TwoFig[6]{
  \sbox\IBoxA{\includegraphics[width=0.45\textwidth]{#1}}
  \sbox\IBoxB{\includegraphics[width=0.45\textwidth]{#4}}%
  \ifdim\ht\IBoxA>\ht\IBoxB
    \setlength\IHeight{\ht\IBoxB}%
  \else\setlength\IHeight{\ht\IBoxA}\fi
  \begin{figure}[!htb]
  \minipage[t]{0.45\textwidth}\centering
  \includegraphics[height=\IHeight]{#1}
  \caption{#2}\label{#3}
  \endminipage\hfill
  \minipage[t]{0.45\textwidth}\centering
  \includegraphics[height=\IHeight]{#4}
  \caption{#5}\label{#6}
  \endminipage 
  \end{figure}%
}

\title{Deep Generative Models, Synthetic Tabular Data, and Differential Privacy: An Overview and Synthesis}
\author{Conor Hassan\footnote{\url{conordaniel.hassan@hdr.qut.edu.au}} $\textsuperscript{ 1,2 }$, Robert Salomone$\textsuperscript{ 1,3}$, and Kerrie Mengersen$\textsuperscript{ 1,2}$ \\ {\small \textsuperscript{1} Centre for Data Science, Queensland University of Technology} \\ 
{\small \textsuperscript{2} School of Mathematical Sciences, Queensland University of Technology}  \\ 
{\small \textsuperscript{3} School of Computer Science, Queensland University of Technology}  \\
}
\date{}
\begin{document}
\maketitle
\begin{abstract}
This article provides a comprehensive synthesis of the recent developments in synthetic data generation via deep generative models, focusing on tabular datasets. We specifically outline the importance of synthetic data generation in the context of privacy-sensitive data. Additionally, we highlight the advantages of using deep generative models over other methods and provide a detailed explanation of the underlying concepts, including unsupervised learning, neural networks, and generative models. The paper covers the challenges and considerations involved in using deep generative models for tabular datasets, such as data normalization, privacy concerns, and model evaluation. This review provides a valuable resource for researchers and practitioners interested in synthetic data generation and its applications.
\end{abstract}

\section{Introduction}
This work considers the task of producing  {\em synthetic} data that reflects the properties of an observed \emph{tabular} data set. Tabular data refers to data organized in rows and columns, akin to a spreadsheet or a relational database table. Each row in the table represents an observation or instance, assumed to be independent, while columns correspond to the attributes or features associated with each observation.
Generating synthetic data that can mimic real data sets in various scenarios enables data custodians to make insights into their data more accessible. Creating realistic synthetic data sets provides access to representative data for building probabilistic models and a means for testing any developed approaches intended for later use on the original data. Additionally, it enables showing or open sourcing of data for broader exploration, application building, and methodological development. 

A crucial aspect of any synthetic data endeavor is to ensure the preservation of the privacy of the original data set. The reason is that data custodians may have large data sets containing confidential, commercially sensitive, or personal information that they are unwilling to release or share due to privacy concerns. Examples of such data sets include health data \citep{buczak2010data, walonoski2018synthea, dankar2013practicing} and survey data \citep{nowok2016synthpop, burgard2017synthetic, abowd2008protective, dwork2019differential}. 

This paper considers {\em generative}, model-based approaches for generating synthetic data. Such approaches are characterized by fitting a probabilistic model to the original data set and generating the synthetic data set by simulating new data from the fitted model. The process of fitting such models can be done in a manner that satisfies the concept of {\em differential privacy} \citep{dwork2014algorithmic}, which provides mathematical guarantees for the privacy of individual observations in the original data set. 

Any model-based approach to synthetic data generation requires a highly flexible class of models that can effectively approximate the complexity of real-world datasets. \emph{Deep generative models} \citep{bondtaylor2021deep, tomczak2022deep} are a class of methods that leverage the expressive power of modern deep learning techniques \citep{bengio2012deep} to model the joint distribution of all variables obtained from a data-generating process, while simultaneously allowing for easy simulation from the fit model.  These models are probabilistic and use parametrizable non-linear functions (neural networks) to obtain high levels of flexibility in the distributions they can fit. The term {\em deep} refers to using neural networks, while {\em generative} refers to the ability to simulate synthetic samples from the fitted model. Deep generative models can also be used in assessing the likelihood of new observations, the labelling of observations, and representation learning \citep{bengio2013representation}. They are ``black-box'' {\em general-purpose} approaches that are widely applicable and do not require domain-specific knowledge, e.g., mathematical modeling of phenomena or positing specific associations between variables as is standard in traditional statistical models. For the above reasons, synthetic data generation via deep generative models holds considerable promise, and research interest in this direction has rapidly grown. 

This work aims to provide a cohesive, top-to-bottom primer on the above topics and some of their recent developments. 
The intended audience is those familiar with statistical modelling but who wish to learn more about the technical aspects relating the generative models, synthetic data, differential privacy, and the intersection of those topics to understand their somewhat disjoint literature. To that end, the contributions of this paper are:
\begin{enumerate}
\item A concise yet cohesive overview of different approaches for constructing (and fitting) flexible classes of parametrized probability distributions to simulate synthetic data; 
\item An in-depth examination of the recent developments in deep generative models specifically for addressing the challenges associated with tabular datasets (as opposed to other data more commonly considered in the generative modelling literature such as images);
\item A discussion of critical works relating to the practical considerations surrounding the process of using deep generative models for synthetic data generation, including the methods involved, adaptations for different variable types, data normalization, privacy considerations, and evaluation of the generated synthetic data; and 
\item An introduction to the concept of differential privacy, and an overview of its developments leading to the widely-applicable \textit{differentially-private stochastic gradient descent} (DP-SGD) method \citep{abadi2016deep}, as well as the alternative Private Aggregation of Teacher Ensembles \citep{papernot2017semisupervised} approach, which are the two main approaches used in differentially-private generative models for synthetic tabular data; an overview of recent such approaches in the literature is also provided.
\end{enumerate}

The structure of the paper is as follows. Section \ref{nn_primer} provides a brief introduction to \emph{neural networks} that form the foundation of deep generative models. Section \ref{generative_models} introduces deep generative {\em modeling} approaches relevant to synthetic tabular data generation. Section \ref{generative_model_inference} outlines the {\em inference} (model-fitting) algorithms used for popular generative modelling approaches. Section \ref{tabular_generative_modeling} discusses adaptions of deep generative models for \emph{tabular} datasets. Section \ref{synthetic_data_evaluation} covers methods for evaluating the high-dimensional synthetic datasets and the models that generate them. Section \ref{differential_privacy_ml} presents an overview of \emph{differential privacy} and its implementation in machine learning models. 
Section \ref{discussion} concludes the paper.

\section{Primer on Neural Networks}\label{nn_primer}
\emph{Neural networks} (NNs) \citep{haykin2004comprehensive} are a type of non-linear parametric function composed of repeated applications of an affine transformation and an elementwise non-linear function. A common use is to approximate complicated functions commonly used in regression and classification tasks. For this article, it is most beneficial to consider neural networks as flexible parameterized functions useful as modular model components.

For simplicity, we focus on feedforward neural networks, also called {\em multilayer perceptrons} (MLPs). For settings involving spatial or temporal dependencies, architectures such as convolutional neural networks (CNNs) \citep{lecun1989backpropagation}, recurrent neural networks (RNNs) \citep{hopfield1986computing}, or transformers  \citep{vaswani2017attention} are potentially more appropriate. For a comprehensive overview of NNs, see \cite{goodfellow2016deep}. 

Here, we describe the construction of an MLP of arbitrary {\em depth}. The input to the NN is $\v x \in \mathbb{R}^D$, and the output is $\mu \in \mathbb{R}^M$. The simplest example of a neural network is the affine function $g_0(\v x) = \m W_1 \v x + \v b_1$, where the parameters are $\m W_1 \in \bb R^{M \times D}$ and $\v b_1 \in \bb R^M$. The notations $\m W_1$ and $\v b_1$ refer to {\em weights} and {\em biases}, as $\m W_1$ weights the input vector, relative to the output vector, and $\v b_1$ biases the resulting output.

A more flexible class of function can be created by applying an elementwise non-linear function $\sigma(\cdot)$ to an affine function and taking another affine combination of the resulting output: $g_1(\v x) = \m W_2 (\sigma(\m W_1 \v x + \v b_1)) + \v b_2$.
This is an example of a {\em single} hidden layer neural network, as it computes an intermediate \emph{hidden} vector $\v h_1(\v x) := \sigma(\m W_1 \v x + \v b_1)$ that we then apply an affine function too. Note that in this setup we may have $\m W_1 \in \bb R^{H_1 \times D}$, $\m W_2 \in \bb R^{M \times H_1}$, $\v b_1 \in \bb R^{H_1}$ and $\v b_2 \in \bb R^M$ for an arbitrary dimension of the hidden vector $H \in \bb N$. For an activation function, the \emph{rectified linear unit} (ReLU) \citep{nair2010rectified}, $\text{ReLU}( x)=\max\big\{0, x\big\}$, is commonly used. For other choices, see, e.g.,  \cite{nwankpa2018activation}. By extension, one can construct parameterized functions with an arbitrary number of hidden layers by continuing the application of activation functions and affine transformations in the above manner. For example, a function involving two hidden layers would take the form $g_2(\v x) =  \m W_3 \sigma(\m W_2 \v h_1(\v x) + \v b_2) + \v b_3$, where now $\m W_3 \in \bb R^{M \times H_2}$, $\v b_3 \in \bb R^{M}$, $\m W_2 \in \bb R^{H_2 \times H_1}$, and $\v b_2 \in \bb R^{H_2}$. The activation functions used in each layer are often kept the same in practice. 

Neural networks are parametric classes of functions. Parameter estimation is often via minimizing a {\em loss} function. The simplest example is using neural network functions as part of a regression model. For regression with a real-valued response $y$ and covariates $\v x$, a commonly used objective is to minimize the {\em squared loss} function over the $N$ observations $\mathcal{L}(\v \theta) = \sum_{i=1}^N \left(y_i - g_{\v \theta}(\v x_i)\right)^2$, where $g_{\v \theta}$ is some neural network function parameterized by $\v \theta$ (which contains the weight matrices and bias vectors). Minimizing the least-squares loss function is equivalent to maximizing the Gaussian log-likelihood with an arbitrary variance parameter over a collection of observations that are considered independent.

If a neural network $g$ is required to map outputs to a constrained subset of $\bb R^M$, an appropriate function can be applied following the final affine transformation as a final part of $g$. For example, letting $\v v$ represent the output of the final affine transformation for neural networks used as part of classification models, i.e., multinomial regression, the softmax function, $\mu_k = \exp(v_k)/(\sum_{k=1}^K \exp(v_k))$ is applied. With such an output, it is typical to set the objective as minimizing the \emph{multiclass cross-entropy} loss \citep{cox1958regression},
\begin{align*}
    \mathcal{L}(\v \theta) = -\frac{1}{N}\sum_{i=1}^N \sum_{k=1}^K y_{k}^{(i)}\log(\mu_{k}^{(i)}). 
\end{align*}
In the above equation, $\v y_i= (y_1^{(i)}, \ldots, y_K^{(i)})$ represents the {\em one-hot encoded label} of the $i^{th}$ observation for a possible $K$ classes. In a one-hot encoding, each class label is converted to a binary vector, where $y_{k}^{(i)} = 1$ if observation $i$ is of class $k$, and all other elements are set to $0$. The term $\mu_{k}^{(i)}$ represents the estimate from the NN function of the probability that $y_i^k=1$. The minimization task is equivalent to the maximum likelihood estimation task of the corresponding multinomial regression model. If no hidden layers exist in $g$, and $y$ is a binary response, the model reduces to logistic regression. 

The above examples are special cases of nonlinear extensions of generalized linear models (GLMs) \citep{nelder1972glm}.
GLMs model the response variable as coming from a chosen \emph{exponential family distribution} \citep{pitman1936sufficient, darmois1935surles, koopman1936distributions}, where the conditional mean $\mathbb{E}[Y\big |\v X= \v x] = g(\mu)$ for some specified {\em inverse-link} function $g$. The principled combination of a loss function and the judicious use of neural networks is a recurring theme throughout this article. However, loss minimization does not always correspond to likelihood maximization.

A more sophisticated example of using neural network functions is \textit{autoencoders} \citep{ballard1987modular}. Such an approach uses two non-linear functions to generate and learn a low-dimensional representation of the observations. The encoder network, $f_\text{encoder}(\v x):\mathbb{R}^D\rightarrow \mathbb{R}^L$, takes as input data observations $\v x$ and outputs a latent representation $\v z$. The decoder network, $f_\text{decoder}(\v z):\mathbb{R}^L \rightarrow\mathbb{R}^D$, takes $\v z$ as input and outputs a reconstructed value $\tilde{\v x}$. The autoencoder model is $f_{\rm decoder} \circ f_{\rm encoder}$, and is trained by minimizing $\mathcal{L}(\v x, \tilde{\v x}) := ||\v x - \tilde{\v x} ||^2$, the so-called {\em reconstruction loss}. Such a procedure is a form of dimensionality reduction as the dimension of the latent representation is often chosen such that $L << D$, meaning that autoencoders implicitly learns a lower-dimensional embedding of an input $\v x$ via $f_{\rm encoder}$ that preserves as much information as possible as defined by the ability to reconstruct the vector via $f_{\rm decoder}$.  

Neural networks have many different forms and applications. However, despite differences in architecture and/or loss functions used, they are typically trained (fit) using the same techniques. \emph{Gradient descent} methods \citep{robbins1951stochastic, rumelhart1986learning}, where the gradients are calculated via automatic differentiation (AD), are typically used to optimize any objective function involving a neural network (or multiple networks) with respect to their parameters. For an overview of AD, see \cite{baydin2018automatic}. As all computations are matrix or elementwise operations, training is easy to parallelize and is suitable for graphical processing units (GPUs). Gradient descent is an iterative algorithm where at each iteration, the set of parameters is updated in the negative direction of their gradient with respect to the objective function, 
\begin{equation*}
\v\theta^* = \v\theta - \eta \nabla_{\v\theta}\mathcal{L}(\v\theta), 
\end{equation*}
where $\nabla_{\v\theta}\mathcal{L}$ is the gradient of some {\em objective function} with respect to the parameters $\v\theta$, and the \emph{learning rate} $\eta$ is a hyperparameter. The algorithm iterates until convergence. In practice, \emph{stochastic gradient descent} (SGD) is more commonly used. Here, the gradient at each iteration is calculated using a small subset of the observations, a \emph{subsample}. Subsampling increases computational efficiency and reduces overfitting because calculating the gradient estimator with different subsamples at each iteration reduces the chance of getting stuck in an undesirable local minima \citep{keskar2017improving}. Several variants exist that adaptively choose the learning rate $\eta$ online; for an overview, see \cite{ruder_2016_gradient_descent}.

\section{Deep Generative Models}\label{generative_models}
\subsection{Overview}\label{dgm_overview}
Deep generative models (DGMs) are models that define a data-generating process through the combination of probabilistic modelling and deep neural networks. We consider a common framework involving data observations $\v x\in\mathcal{X}$,  latent variables $\v z\in\mathcal{Z}$, and model parameters $\v \theta \in \Theta$. The different types of DGMs are defined by how neural networks model the relationship between latent variables and data and can be either {\em prescribed}, having some available form of $p_{\v \theta}(\v x)$, or {\em implicit}, where the associated density function is not available in closed form. The probabilistic modelling approach specifies a joint probability model for $\v z$ and $\v x$. Then, the generation of new samples from a model with fixed parameters involves sampling latent variables $\v z \sim p_{\v\theta}(\v z)$ from a prior distribution 
 $p_{\v\theta}(\v z)$ and subsequently generating data $\v x \sim p_{\v\theta}(\v x|\v z)$ from the conditional distribution $p_{\v\theta}(\v x|\v z)$.
 Section \ref{latent_variable_models} describes methods based on the above approach that allow for the construction of flexible families of marginal distributions $p_{\v\theta}(\v x)$. Such models can be extensions of classical approaches such as mixture models and factor analysis. 
 
Approaches also exist that employ a \textit{deterministic} mapping from the latent space to the data space. Here, flexible distributions are created by applying some deterministic function $T_{\v \theta}:\mathcal{Z}\rightarrow\mathcal{X}$ with parameters $\v \theta$ that maps latent variables to observations. Section \ref{flow_based} introduces \textit{flow-based} models, which often employ neural networks as building blocks in the construction of bijective $T_{\v \theta}$ in a manner facilitating the explicit computation of the resulting log-likelihood function. Section \ref{gan} introduces {\em Generative Adversarial Networks}, which allow for more general classes of $T_{\v \theta}$ (e.g., using neural networks themselves) and overcome the need for an explicit likelihood by using a surrogate objective based on a concept called {\em adversarial learning}.

Section \ref{subsect:conditionalDGMs} concludes with a discussion regarding {\em conditional} extensions of the above approaches, extending the ideas discussed from the setting of modelling some general joint distribution $p(\v x)$ to the setting of modelling some conditional distribution $p(\v x | \v c)$ for some {\em conditioning} vector $\v c$.
 
\subsection{Latent Variable Models}\label{latent_variable_models}
In the context of generative models, latent variable models are a type of \emph{unsupervised learning} or \emph{density estimation} that describe a data-generating process, $ \v z \sim p_{\v\theta}(\v z)$, $\v x | \v z \sim p_{\v\theta}(\v x | \v z)$. Observations simulated from the generative model are {\em synthetic} data observations. Latent variables $\v z$ are unobserved variables; their primary use in our context is to aid in creating more flexible models as the induced marginal density over the observed variables involves integrating over $\v z$. Latent variable models with a low-dimensional latent vector compared to the dimensionality of the observed variables are often also helpful in extracting low-dimensional hidden structure from data, e.g., by modelling a lower-dimensional subspace (manifold) that the data lie primarily around.

Examples of latent variable models commonly used in the statistics literature are mixture models \citep{mengersen2011mixtures} and factor analysis models \citep{bartholomew2011latent}. Mixture models use latent variables to represent unobserved \emph{mixture components} within the population from which the dataset was drawn. A Gaussian mixture model (GMM) \citep{duda1973pattern, mclachlan1988mixture} is a type of mixture model that assumes each observation has a single latent variable that takes one of $K$ values, representing the $K$ mixture components. The base distribution for each latent variable is the categorical distribution $\v z \sim \mbox{Categorical}(\v \pi)$, and the conditional distribution $\v x | \v z \sim p_{\v\theta}(\v x | \v z )$ is Gaussian, i.e., $p(\v x | \v z = k) = \mathcal{N}(\v \mu_k, \m \Sigma_k)$,  giving the parameters $\v\theta = (\v \pi, \v \mu, \v \Sigma)^\top$, where $\v\pi = (\pi_i, \ldots, \pi_K)^\top$, $\v \mu = (\v\mu_1, \ldots, \v\mu_K)^\top$, and $\v \Sigma = (\m \Sigma_1, \ldots, \m \Sigma_K)^\top$. The combination of this base and conditional distribution defines a GMM.

Another probabilistic model is a \emph{factor analysis} model. Here, Each observation $\v x \in \bb R^{d}$ has an associated lower-dimensional latent vector $\v z\in\mathbb{R}^L$. The simplest base density for $\v z$ is $\v z \sim \mathcal{N}(\v\mu_0, \m\Sigma_0)$. Suppose observations are continuous variables $\v x \in\mathbb{R}^D$. In that case, the likelihood can be Gaussian with a mean vector that is an affine transform of the latent variables $\v z$, given by $\v x | \v z \sim \mathcal{N}(\m W\v z + \v\mu, \m\Psi)$, where $\m W\in\mathbb{R}^{D\times L}$ is called the \emph{factor loading} matrix, $\v \mu \in \mathbb{R}^D$, and $\Psi\in\mathbb{R}^{D\times D}$. Note that $\m W$ implicitly defines a linear subspace of $\bb R^D$ of dimension $L$. If $\m\Psi = \sigma^2 \m I$, then the factor analysis model is equivalent to probabilistic principal component analysis \citep{tipping1999probabilistic}. A mixture of factor analyzers (MFA) \citep{ghahramani1996algorithm} combines the latent representation of mixture models and factor analysis models. The generative model for MFA is to draw a mixture component $k \sim\mbox{Categorical}(\v\pi)$, a latent vector $\v z \sim\mathcal{N}(\v 0, \mbox{I}_L)$, 
 and then generate data through the conditional distribution $\v x | k, \v z\sim \mathcal{N}(\v\mu_k+\m W_k\v z, \m\Psi)$.
 \emph{Deep} latent variable models (DLVMs) are extensions of the model class above that have a hierarchical latent variable structure and/or non-linear aspects. 

Deep latent Gaussian models (DLGMs) \citep{rezende2014stochastic} is a prevalent type of deep latent variable model that employs neural networks. The generative process of a single-layer DLGM is to draw a latent variable $\v z \sim \mathcal{N}(\v 0, \m I_L)$ and then draw from the conditional $\v x | \v z\sim \mathcal{N}(\v\mu_{\v\theta}(\v z), {\rm diag}(\v \sigma_{\v \theta}(\v z))\m I_D)$ where $\v\mu_{\v \theta}(\cdot)$ and $\v\sigma_{\v \theta}(\cdot)$ are obtained via joint output of a regular neural network termed the {\em decoder network}. Note that such a construction yields a continuous (infinite) Gaussian mixture model and hence can be expected to be potentially very flexible. The example just discussed is an example with one hidden layer of latent variables. Extensions are possible that involve several layers of latent variables, each parametrized by the output of some neural network taking the previous layer as input.

More generally, the prior distribution of the latent variables $p_{\v\theta}(\v z)$ and the conditional distribution of $p_{\v\theta}(\v x | \v z)$ can be arbitrary. Deep exponential families (DEFs) \citep{ranganath2015deep} generalize DLGMs to models where the latent variables at each layer can come from any exponential family distribution. Sigmoid belief networks \citep{neal1990learning} are an early example of a DEF with layers of binary latent variables. As discussed in Section \ref{variational_autoencoders}, variational autoencoders (VAEs) \citep{kingma2014autoencoding} are simply DEFs trained with something called an amortized inference scheme. The latter allows for, following training, approximate inferences of $\v z|\v x$ for observations $\v x$ that were not contained in the training set to be obtained.

\subsection{Flow-Based Models}\label{flow_based}
{\em Flow-based models}, or {\em normalizing flows} (NFs) \citep{tabak2010density, tabak2013family, rippel2013high, papamakarios2021normalizing}, are a class of generative models that involve constructing flexible classes of probability distributions via transforming a random vector $\v z$ (taking values on $\bb R^D$) with a simple distribution, such that the distribution of the transformed random vector is known. This is achieved using a function $T_{\v\theta}: \bb R^D \rightarrow \bb R^D$, called a {\em flow}, with parameters $\v\theta$. Provided that for any $\v \theta$, the function $T_{\v \theta}$ is a bijective map that is invertible, differentiable and has a differentiable inverse, one can apply
the classic transformation (change-of-variables) theorem. Specifically, if $\v z \sim p(\v z)$ and $\v x = T_{\v \theta}(\v z)$, then it is known that $\v x$ has the density
\begin{align}\label{normalizing_flow_likelihood}
p_{\v\theta}(\v x) = p_{\v z}\big (T_{\v\theta}^{-1}(\v x)\big )\big{|}{\rm  det }J_{T_{\v \theta}^{-1}}(\v x)\big{|},
\end{align}
where the notation $J_{{T}}(\v x)$ represents the Jacobian matrix of a function $T$ evaluated at the point $\v x$. The construction of flexible classes of distribution is thus equivalent to one of defining a flexible parametrized class of transforms that satisfies the mentioned constraints. The distribution $p(\v z)$ is called the {\em base distribution}.
The construction above allows for both easy simulation of $\v x$ and evaluation of the likelihood $p_{\v\theta}(\v x)$ for parameter estimation using maximum likelihood estimation.

The familiar multivariate Gaussian distribution has a representation as a flow-based model. Here, taking $\v z \sim \mathcal{N}(\v 0, \m I_D)$ and $T_{\v\theta}(\v z) = \v \mu + \m L \v z$ yields that $\v x \sim \mathcal{N}(\v \mu, \m \Sigma)$ where $\m \Sigma = \m L\m L^\top$. In other words, multivariate normal distributions arise from applying an {\em affine} flow to a standard multivariate normal base distribution. In the same way that neural networks can be seen as non-linear regression models, constructing flexible bijective transformations with neural networks applied to a base multivariate normal can be seen as a nonlinear extension of the multivariate normal distribution. 

In more sophisticated flow-based models, the transformation is the composition of a sequence of individual transformations called {\em flow layers} (multiple subtransformations are used to increase model flexibility). That is, $T_{\v\theta}$ is the composition of flow layers $T_1,\ldots, T_K$, each of which typically have their own subset of parameters contained in $\v \theta$, and $T_{\v \theta}(\v z) = T_{K}\circ T_{K-1}\circ\cdots\circ T_2\circ T_1(\v z)$. Two identities for the composition of $K$ functions make such an approach elegant. Firstly, the inverse of a composition of functions is the inverse of each function iterated backwards,  
\begin{equation}
T_{\v \theta}^{-1}(\v x) = (T_K \circ \cdots\circ T_1)^{-1}(\v x) = T_1^{-1}\circ \cdots \circ T_K^{-1}(\v x).  
\end{equation}
Second, by defining $\v z^{(k)}=  (T_k \circ \cdots\circ T_1)^{-1}(\v x) $ for $k>1$ and simply $T_1^{-1}(\v x)$ for $k=1$, we have the identity
\begin{equation}
\mbox{det }J_{T_1^{-1}\circ \cdots \circ T_K^{-1}}(\v x) = \mbox{det }J_{T_1^{-1}}(\v z^{(1)}) \cdots \mbox{det }J_{T_{K-1}^{-1}}(\v z^{(K-1)}) \cdot \mbox{det }J_{T_K^{-1}}(\v x).
\end{equation}

Hence, the computational cost of computing the (log) likelihood scales linearly in the number of transformations in the composition. As the computation of matrix determinants is generally of complexity $\mathcal{O}(D^3)$, a common strategy is to design flow layers that yield a lower-triangular Jacobian matrix. In the latter case, the matrix determinant is simply the product of diagonal entries and reduces to $\mathcal{O}(D)$ complexity.

We now provide some examples of commonly used flow-based models. The approaches are described in terms of a single flow layer for ease of presentation. However, more generally, several such layers are composed in practice. An example of such a flow layer is the {\em masked autoregressive flow} (MAF) layer \citep{papamakarios2017masked}. Let $\v \mu_{\v \theta}: \bb R^{D} \to \bb R^D$ and $\v \sigma_{\v \theta}: \bb R^{D} \to \bb R^D$ be neural network functions such that the $i$-th output of each depends only on the first $i-1$ inputs. Such functions are constructed in practice by masking the parameters of a standard neural network appropriately; see \cite{germain2015made} for details. Here, we define $\v v_{1:m}$ to represent the vector containing the first $m$ elements of a vector $\v v$. Then, a MAF layer is defined as
 \[x_i =  \mu_{\v \theta}(\v x_{1:i-1})_i 
 + {\exp \sigma_{\v\theta}(\v x_{1:i-1})_i}\, z_i, \quad i = 1,\ldots, D.\]

 In practice, an autoregressive neural network that outputs both $\v \mu_\theta(\cdot)$ and $\v \sigma_{\v \theta}(\cdot)$ each individually respecting the autoregressive property can be employed. Noting that $z_i = \frac{x_i - \mu_{\v\theta}(\v x_{1:i-1})_i}{\exp \sigma_{\v\theta}(\v x_{1:i-1})_i}$ reveals that all elements in $\v z$ can be computed in parallel for a given $\v x$. Such a property is useful for fast density evaluation, and hence maximum-likelihood training, on a set of observations. However, for a fitted masked autoregressive flow model, generating new synthetic observations is a {\em sequential} operation. If the main objective of the generative model is to generate data in real-time or for repeated use cases, such that the efficiency when generating new data is more important than the speed of training, then the inverse of MAF layers, described shortly, may be preferred.

{\em Inverse autoregressive} flow (IAF) layers  \citep{kingma2016improved} are normalizing flow layers that allow efficient generation of synthetic data in parallel. In an IAF layer, the function instead takes the form, 
 \[x_i =  \mu_{\v \theta}(\v z_{1:i-1})_i 
 + {\exp \sigma_{\v\theta}(\v z_{1:i-1})_i}\, z_i, \quad i = 1,\ldots, D,\]
 where the neural network functions are defined similarly to before but instead now depend on elements of $\v z$ as opposed to $\v x$. The advantage of IAF layers is that generation is a parallel operation, and the inverse is now sequential. The affine flow seen in the earlier multivariate Gaussian example is an example of an IAF in the case where the matrix $\m L$ is lower triangular.
 
In contrast to a fully autoregressive structure, {\em real non-volume preserving} (RealNVP) flows \citep{dinh2016density} use learnable transformations called {\em affine coupling layers}. Here, $\v z$ is partitioned into $\v z_1$ and $\v z_2$. Then,  $\v x =  (\v z_A, \mu_{\v \theta}(\v z_A)  + \exp \sigma_{\v\theta}(\v z_{A}) \odot \v z_B)$, where $\mu_{\v \theta}(\cdot)$ and $\v \sigma_{\theta}(\cdot)$ are neural network functions of appropriate dimension (again, often obtained as a joint output of a single neural network in practice), and $\odot$ refers to the elementwise product.
To ensure that all variables can be flexibly transformed, layers of any flow type are often interspersed with a {\em permutation layer}, for example, a function that reverses the order of variables. Like autoregressive flows, real NVP layers necessarily have lower-triangular Jacobian matrices. However, they enjoy parallel computation in both directions. 

MAF, IAF, and Real NVP can each be viewed as applying a collection of univariate {\em affine} transformations to each $\v z_i$ for which the scale and shift parameters are determined by some potentially non-linear function of some subset of the other parameters. A natural extension is to replace such affine transformations with univariate \emph{monotonic splines}, which capture a broader function class while being easily invertible. A monotonic spline is a piecewise function consisting of $S$ segments, where each segment is easy to invert. Given a set of $S+1$ locations $v_{0}, v_{1}, \ldots, v_{S}$, the transformation is a simple monotonic function such as a low degree polynomial within each interval $[v_{(j-1)}, v_{j}]$, for $j\geq 1$. The only constraint is that the segments must meet at their endpoints. Outside the range of $[v_{0}, v_{S}]$, the transformation can default to a simple function such as the identity function. Spline-based transformations are distinguished by the type of spline used. Some of the options that have been explored so far, in order of increasing flexibility, are linear and quadratic splines \citep{muller2019neural}, cubic splines \citep{durkan2019cubic}, linear-rational splines \citep{dolatabadi2020invertible}, and rational-quadratic splines \citep{durkan2019neural}. Implementations of many of the aforementioned approaches can be found in the Python package {\sf nflows} \citep{stimper2023normflows}.  

There are many other possible types of flow layers. {\em Residual flows} \citep{chen2019residual} use residual connections, popular within residual networks \citep{he2016deep}, with additional parameter constraints to ensure invertibility within the transformation. {\em Continuous-time flows} such as neural ODEs \citep{chen2018neural} and FFJORD \citep{grathwohl2018ffjord} use a differential equation to define flow layers and consider a continuous analogue of the change-of-variable formula. For a comprehensive survey of flow-based models, including a discussion and overview of methods in the case of discrete distributions, see \cite{papamakarios2021normalizing}.

\subsection{Generative Adversarial Networks}\label{gan}
Generative adversarial nets (GANs) \citep{goodfellow2014generative} are a class of deep generative models that have enjoyed considerable popularity in recent years. 
Arguably, this is due to their simple construction and impressive performance on various tasks, particularly their original goal of being a generative model that aims to produce plausible-looking images. GANs are now widely used for tabular data and other data sets.

The generative approach underlying GAN models are very similar to that of flow-based models: draw latent variables $\v z$ from some base distribution $p(\v z)$, and then output $\v x = T_{\v \theta}(\v z)$. The fundamental difference is that GANs use an alternative loss function in training that does not involve likelihood and therefore places significantly fewer restrictions on the function $T_{\v \theta}$, which in the GAN literature is referred to as the {\em generator}. The function $T_{\v \theta}$ may be any function and is typically a neural network, mapping from $L$-dimensional space to $D$-dimensional space. Unlike flow-based models, $T_{\v \theta}$ need not be invertible, and $L$ need not equal $D$. However, in contrast to flow-based models, GANs do not necessarily and typically do not yield a tractable likelihood. 

Adversarial learning involves training an auxiliary network called the \emph{discriminator} $d_{\v\lambda}:\mathcal{X} \to \mathcal{C}$. The discriminator is trained to quantify the extent to which a given input is a sample from the true data-generating process or simply a sample simulated from the generator. Commonly, a neural network suited to binary classification problems is used, i.e., $\mathcal{C} = (0,1)$, as $d_{\lambda}$ outputs a predicted probability that the $\v x$ given as input comes from the true probability distribution). The objective in adversarial learning is 
\begin{equation} \label{gan_optimal}
    \text{arg}\min_{\v{\theta} \in \v{\Theta}} \left\{\max_{\v{\lambda} \in \v{\Lambda}} L(\v{\theta}, \v{\lambda}) \right\}.
\end{equation} 
where $L$ is some loss function. When training the discriminator as a binary classifier, the loss function for a single observation in the original formulation is the cross-entropy loss:
\begin{equation}\label{gan_loss_function}
    L(\v{\theta}, \v{\lambda})=\log d_{\v\lambda}(\v{x}) + \mathbb{E}_{\v z \sim p({\v z})}\log(1 - d_{\v\lambda}(T_{\v \theta}(\v z))),
\end{equation}
where the final term in the sum is an expectation as it corresponds to all possible outputs of the generator. An average of such functions is used as the overall loss $\mathcal{L}(\v \theta, \v \lambda)$. The fundamental idea underlying adversarial learning is that by simultaneously training the generator to ``fool'' the discriminator into thinking its generated samples are real while also training the discriminator to identify generated samples better. 

 Most notably, constructing generators outputting categorical variables or a mixture of continuous and categorical variables is straightforward in the adversarial training setting. There is also a considerably large amount of variants in terms of loss functions, but the essential idea remains similar. An approach of note is FlowGAN \citep{grover2018flow}, which combines maximum likelihood estimation and adversarial learning by implementing the generator as a normalizing flow and discusses the merits of each approach. 
 A common issue in training GANs is a phenomenon called \emph{mode collapse}. Mode collapse arises during training, resulting in a generator that creates samples that are plausible to the discriminator but not representative of the entire distribution of the observed data. Many papers have proposed alternate objective functions to avoid mode collapse. A popular example is the Wasserstein GAN (WGAN) \citep{arjovsky2017wasserstein}. For a comprehensive recent survey of GAN-based approaches, see \cite{gui2021review}. 
 
\subsection{A Note on Conditional Variants of DGMs} \label{subsect:conditionalDGMs}
It is worth noting that the models discussed in this section focussed on either explicitly or implicitly modeling flexible classes of parametrized distributions $p_{\v \theta}(\v x)$. All ideas discussed readily extend to the case of modelling instead {\em conditional} distributions, i.e., $p(\v x|\v c)$ for some input $\v c$, often called the {\em context} vector. For example, conditional normalizing flows \citep{winkler2019learning} and conditional GANs \citep{mirza2014conditional} are both straightforward extensions. Such settings are sometimes called {\em regression density estimation} in the statistics literature \citep{nott2012regression}. An early approach where the response density is a mixture model with parameters given by the output of a neural network that is given $\v c$ as input is mixture density networks \citep{bishop1994mixture}.

In principle, it is possible to create a model using two or more types of generative models, each for a part of the data set. For example, the continuous variables $\v x_1$ may be fit using a normalizing flow to obtain $p_{\v \theta_1}(\v x_1)$, and then discrete variables $\v x_2$ can be fitted using a conditional GAN with $\v x_1$ as a context vector, thus learning $p_{\v \theta_2}(\v x_2|\v x_1)$. Together, the two models define a joint distribution $p_{\v \theta}(\v x_1,\v x_2) = p_{\v \theta_1}(\v x_1)p_{\v \theta_2}(\v x_2|\v x_1)$ for which generation is straightforward. To our knowledge, such approaches have yet to be considered in the literature, and the possible advantages of such an approach (if any) appear unexplored.

\section{Inference for DGMs}\label{generative_model_inference}
In this section, we describe inference (model fitting) methods for the different classes of DGMs introduced in the last section. 
Additionally, we introduce variational autoencoders (VAEs) as a generative model resulting from deep latent variable models trained by amortized variational inference and discuss the relationship between modern variational inference techniques and the expectation-maximization (EM) algorithm. For a comprehensive overview of stochastic (gradient) optimization for machine learning, see \cite{mohamed2020monte}. See \cite{blei2017variational, zhang2018advances} for an overview of variational inference. For a comprehensive overview of the EM algorithm and its variants, see \cite{mclachlan2007algorithm}. 
Throughout this section, we write $\v x_k$ as the $k$th observation contained in a tabular data set with $N$ observations, i.e., $\v x_k$ represents the $k$th row of a data set. 

\subsection{Inference for Deep Latent Variable Models}\label{evidence_lower_bound}

In an idealized scenario, deep latent variable models would maximize the {\em log-likelihood} over the observed data, which involves integrating over the unseen latent variables,
\[\ell(\v \theta) = \sum_{k=1}^N \log p_{\v \theta}(\v x_k) = \sum_{k=1}^N \log  \int p_{\v \theta}(\v x_k, \v z_k)d\v z_k.\] 

However, the integrals in the above objective function render direct optimization via gradient descent infeasible. Subsequently, an approach called {\em variational inference} is used, which is closely related to the expectation-maximization (EM) algorithm used to fit latent variable models such as GMMs and probabilistic PCA. We discuss this relationship further shortly. In the following presentation, we write $\v x$ to represent all observations and $\v z$ for all latent variables.

The idea of variational inference, when applied to latent variable models, is to train an approximation $q_{\v \phi}(\v z)$ for each observation, called either a {\em variational approximation} or the {\em variational posterior}, of $p_{\v \theta}(\v z | \v x)$. Fitting the variational approximation is accomplished by optimizing over $\v \phi \in \Phi$, called the {\em variational parameters}, over some distributional family. For example, if $q_{\v \phi}$ is comprised of all distributions corresponding to $D$ independent but not identically distributed Gaussians, then $\v \phi$ would consist of the mean vector $\v \mu$ and the diagonal of the covariance matrix $\v \sigma$. Often, the parameters are chosen so that any optimization occurs on the unconstrained space, i.e., $\log \v\sigma$ in the current example. The variational approximation may be chosen to factorize across individual observations, i.e., $q_{\v \phi}(\v z) = \prod_{k=1}^N q_{\v \phi_k}(\v z_k)$, where $\v \phi_k$ represent the subset of parameters used in the variational approximation of $p_{\v \theta}(\v z_k|\v x_k)$. Such a choice is termed {\em mean-field} variational inference (MFVI). 

For any choice of $\v \theta$ and $\v \phi$,  we have the associated the {\em evidence lower bound} (ELBO): 
\[{\rm ELBO}(\v \theta, \v\phi) := 
\mathbb{E}_{q_{\v \phi}(\v z)}\big[ \log p_{\v \theta}(\v x, \v z) - \log q_{\v \phi}(\v z)\big ].  
\]
As the name suggests, ${\rm ELBO}(\v \theta, \v \phi)$ for an arbitrary $\v \phi$ is a lower bound for the marginal likelihood $\ell(\v\theta)$, also known as {\em evidence}, as it can be shown (see for example, \cite{blei2017variational}) that, 
\begin{equation} \label{eq:EM_identity}\log p_{\v \theta}(\v x) = {\rm ELBO}(\v \theta, \v \phi) - \bb K \bb L(q_{\v \phi}(\v z)||p_{\v \theta}(\v z|\v x)),\end{equation}
where $\bb K \bb L(q||p)$ is the \textit{Kullback-Leibler divergence}. The latter quantity is called the variational gap and is non-negative and zero if and only if $q = p$, so consequentially ${\rm ELBO}(\v \theta, \v \phi)$
is equal to $\ell(\v\theta)$ if and only if $q_{\v \phi}(\v z) \equiv p_{\v \theta}(\v z | \v x)$. Consequentially, the resulting objective is to minimize
\[\mathcal{L}(\v \theta, \v \phi) = - {\rm ELBO}(\v \theta, \v \phi),\]
jointly with respect to $\v \theta$ and $\v \phi$. The reasons are that (1) minimizing the objective ${\cal L}$ with respect to $\v \phi$ is equivalent to minimizing the total variational gap, thus giving a tighter lower bound on the likelihood, and (2) minimizing $\cal{L}$ with respect to $\v \theta$ is equivalent training the generative model to maximize a lower bound on the likelihood as a surrogate for the actual likelihood. As with most DGMs, the optimization can employ stochastic gradient descent methods. However, the objective function includes terms that involve expectations, so we must consider appropriate estimators of such quantities. 

First, we consider the simpler case of the gradient with respect to $\v \theta$. Here, we have the identity 
\[\begin{split}\nabla_{\v \theta} {\rm ELBO}(\v \theta, \v \phi) &= \nabla_{\v \theta}
\mathbb{E}_{\v z \sim q_{\v \phi}(\v z)}\big[ \log p_{\v \theta}(\v x, \v z) - \log q_{\v \phi}(\v z)\big] \\ 
&= \mathbb{E}_{\v z \sim q_{\v \phi}(\v z)}\big[\nabla_{\v \theta}\left(\log p_{\v \theta}(\v x, \v z) - \log q_{\v \phi}(\v z)\right)\big],
\end{split}
\] 
where in the second equality, the gradient is allowed to pass inside the expectation under very mild conditions, as the expectation is not taken with respect to a distribution that depends on $\v \theta$. Consequentially, an unbiased Monte Carlo estimator for the gradient of the ELBO with respect to $\v \theta$ is obtained by drawing $\v z \sim q_{\v \phi}(\v z)$ and returning 
\[\begin{split}\widehat{\nabla}_{\v \theta} {\rm ELBO}(\v \theta, \v \phi)
:= \nabla_{\v \theta}\left(\log p_{\v \theta}(\v x, \v z) - \log q_{\v \phi}(\v z)\right). 
\end{split}
\] 
Such Monte Carlo gradient estimators that permit the interchange of a gradient and expectation operators are often referred to as {\em infinitesimal perturbation analysis} (IPA). See \cite{lecuyer1990unified} for more details.
However, obtaining an estimator for the gradient with respect to $\v \phi$ is not as straightforward, as the expectation involves a distribution that depends on $\v \phi$. A universal approach that works in such settings is to use an approach called the score function (also called the REINFORCE) gradient estimator \citep{williams1992simple}. However, such estimators typically have a sufficiently large variance to make their use untenable without many samples and variance reduction techniques \citep{blackboxVI_ranganath14}. For this reason, REINFORCE estimators are typically avoided in practice.

Fortunately, practical low-variance gradient estimators can be obtained via the {\em reparameterization trick}. Namely, if $\v z \sim q_{\v \phi}(\v z)$ can be generated via setting $\v z = T_{\v \phi}(\v \epsilon)$ where $T_{\v \phi}$ is some function that depends on $\v \phi$ and $\epsilon \sim p(\v \epsilon)$ comes from a distribution that does not depend on the variational parameters $\v \phi$, then we have the identity that
\[\nabla_{\v \phi}{\rm ELBO}(\v\theta, \v \phi) = 
\nabla_{\v \phi} \mathbb{E}_{\v \epsilon \sim p(\v \epsilon)}\big[\log p_{\v \theta}(\v x, T_{\v \phi}(\v \epsilon)) - \log q_{\v \phi}(T_{\v \phi}(\v \epsilon))\big ],  
\]
and hence, infinitesimal perturbation analysis gradient estimators similar to those previously described can be employed, as the reparameterization trick removes the problematic dependency in the expectation. The ensuing gradient estimators, in this context, are sometimes referred to as {\em reparameterization gradient estimators}.  

Estimators of this kind typically have a sufficiently low variance that gradient optimization is often sufficiently stable for practical purposes by taking only a {\em single} Monte Carlo sample. However, reparameterization-gradient approaches require a differentiable $T_{\v \phi}$ so, in general, can not be applied when $q_{\v \phi}$ is a discrete distribution, i.e., the latent variables $\v z$ are discrete, as in mixture models involving categorical latent variables. In that case, methods such as the EM algorithm may be preferred. However, most deep latent variable models in practice employ continuous latent variables.

In terms of obtaining families of distributions that are amenable to the reparametrization trick, note that taking any family for $q_{\v \phi}$ as one arising from a flow-based model (Section \ref{flow_based}), the simplest of which is arguably a multivariate Gaussian immediately,  satisfies the required desiderata by construction. For this reason, normalizing flows are a topic of independent interest in the variational inference literature \citep{rezende2015variational}, as they can be used to construct flexible variational families amenable to the reparameterization trick. 

We have focussed on the ELBO objective. Tighter lower bound objectives for the marginal likelihood exist, such as the importance-weighted bound \citep{burda2015importance}. For directions extending upon the approach of the latter reference, see \cite{doucet2023differentiable} for a survey of how low-variance Monte Carlo estimators of the marginal likelihood can be used to construct tight lower bounds. 

\subsection{Expectation-Maximization Algorithm and its Relationship to VI}

Maximization of the ELBO objective may at first appear as a foreign approach. However, it is worth noting that it is directly related to the expectation-maximization (EM) algorithm \citep{mclachlan2007algorithm} and its variants. For brevity, we draw the connection between variational inference and the standard form of the EM algorithm.  

The EM algorithm is traditionally used to fit latent variable models where the posterior distributions $p_{\v\theta}\big (\v z \big | \v x \big )$ are available in closed form. It involves two steps, the \emph{expectation step} (E-step) and the \emph{maximization step} (M-step). The E-step consists of fixing $\v \theta$ and computing 
$\mathbb{E}_{p_{\v\theta}(\v z | \v x)}\log p_{\v\theta}\big (\v x, \v z\big )$, in the 
M-step, $\v \theta$ is updated to be the solution of
$\arg \max_{\v \theta}\mathbb{E}_{p_{\v \theta} (\v z | \v x)}\log p_{\v \theta}\big( \v x, \v z\big)$. The two steps are repeated until convergence of the expectation calculated in the E-step.

Recalling equation \eqref{eq:EM_identity}, for fixed $\v \theta$, the E-step can be interpreted as using a variational family that includes the true posterior distribution $p_{\v\theta}(\v z | \v x)$ and setting $\v \phi$ equal to the $\v \phi^\star$ such that  $q_{\v \phi^\star}(\v z|\v x) = p_{\v \theta}(\v z|\v x)$. Thus, the ELBO reduces to 
\[{\rm ELBO}(\v \theta, \v \phi^\star) = \mathbb{E}_{\v z \sim p_{\v\theta}(\v z | \v x)}\log p_{\v\theta}\big (\v x, \v z\big),\]
which is precisely the quantity maximized over $\v \theta$ in the M-step. Thus, the EM algorithm can be interpreted as an algorithm that performs block-coordinate ascent on the ELBO objective, where the family of possible distributions $q_{\v \phi}(\v z)$ contains the true posterior distribution $p_{\v\theta}(\v z|\v x)$. The advantage of the EM approach is that it is applicable in the case of discrete latent variables, such as mixture models, including GMMs; such approaches are challenging for stochastic-gradient variational methods. 

However, the disadvantage of the EM approach is that it requires analytically tractable quantities related to the posterior distribution of the latent variables given the observed variables. Several variants relax this requirement (see \cite{mclachlan2007algorithm} for details), but stochastic-gradient methods are typically used for deep latent variable models. The latter is because such approaches allow for amortized inference, as discussed below. 

\subsection{Amortized Variational Inference and Variational Autoencoders}\label{variational_autoencoders}

An important modification of stochastic gradient variational inference for latent variable models is {\em amortized inference}. Here, a single function is trained to take an observation $\v x$ as input and produce the parameters $\v \phi$ of the variational approximation as output. Instead of optimizing $\v \phi$, a neural network called an {\em inference network} $f_{\v \varphi}: \mathcal{X} \to \Phi$, itself parameterized by $\v \varphi$ is trained. Such an approach results in a family of variational approximations of the form $q_{\v \varphi}(\v z) = \prod_{k=1}^N q_{\v \phi_k}(\v z_k)$, where each $\v \phi_k = f_{\varphi}(\v x_k)$. 

The role of the inference network is to learn a mapping from data space directly to the parameters of a variational approximation for the latent variables $\v z$ corresponding to the input observation; the negative ELBO is minimized over the inference network parameters $\v \varphi$. 

There are two advantages to such an approach. Firstly, it is more scalable as subsampling can be used effectively. Subsampling the data and computing the gradients with respect to $\varphi$ results in inference network updates that carry across to {\em all} samples. Additionally, amortized inference is more parameter-efficient than the non-amortized approach, as the number of variational parameters trained remains fixed irrespective of the size of the dataset. Secondly, the trained inference network can approximate the distribution of the latent variables $\v z^\star$ for an observation $\v x^\star$ that is {\em not}
observed during training by simply feeding $\v x^\star$ through the inference network.

The inference network has the interpretation of a stochastic encoder and the original deep latent variable model as a stochastic decoder. 
Combining the two is essentially the probabilistic analogue of the \textit{autoencoder} neural network discussed at the end of Section \ref{nn_primer}. For this reason, deep latent (Gaussian) variable models trained with amortized variational inference schemes as described above are called {\em variational autoencoders} \citep{kingma2014autoencoding, rezende2014stochastic}. 

\subsection{Inference for Flow-Based Models}
The training of a normalizing flow model is simply the minimization of the negative log-likelihood
\[\ell(\v \theta) := -\sum_{k=1}^N \log p_{\v \theta}(\v x_k) = -\left(\sum_{k=1}^N \log p_{\v z}\big (T_{\v\theta}^{-1}(\v x_k)\big)+ \log \big{|}{\rm det }J_{T_{\v \theta}^{-1}}(\v x_k)\big{|}\right),\]
with respect to $\v \theta$. 
Recall that by construction, the transformations are differentiable and invertible, which allows us to calculate the gradients with respect to the parameters easily using automatic differentiation. Hence, the parameters can be fit using gradient descent or a stochastic variant that subsamples the data.

\subsection{Inference for Generative Adversarial Networks}\label{gan_inference}
Recall that in GAN training, the objective function is
\begin{align*}
  \min_{\v \theta} \max_{\v \lambda} L(\v \theta, \v \lambda)= \left(\frac{1}{N}\sum_{k=1}^N \log d_{\v\lambda}(\v x_k)\right) + \mathbb{E}_{\v z \sim p(\v z)} \log\big(1 - d_{\v\lambda}(T_{\v \theta}(\v z))\big).
\end{align*}
As the first term does not depend on the generator parameter $\v \theta$, the GAN objective can be written as minimizing the following two losses simultaneously,
\begin{align*}
    L_d(\v \theta)&=\mathbb{E}_{\v z \sim p(\v z)} \log\big(1 - d_{\v\lambda}(T_{\v\theta}(\v z))\big), \quad \text{ and } \\
    L_g(\v \lambda)&=-\left(\left(\frac{1}{N}\sum_{k=1}^N\log d_{\v \lambda}(\v x_k )\right)-\mathbb{E}_{\v z \sim p(\v z)} \log (1 - d_{\v\lambda}(T_{\v\theta}(\v z))\right).
\end{align*}
GANs are naturally parameterized in the form of the reparametrization trick, enabling gradient estimation. GAN variants commonly have their loss functions decompose into two parts, as above, and are trained similarly.

\section{Generative Modelling for Tabular Data}\label{tabular_generative_modeling}
This section discusses considerations, modifications, and recently-proposed refinements to the models discussed in Section \ref{generative_models} specific to the tabular data setting. Tabular data sets may contain a mixture of {\em continuous}, {\em categorical}, {\em discrete}, and {\em ordinal} variables with complicated characteristics and dependency structure. The challenges of modelling data with such diversity and the properties of different generative model classes necessitate specific adaptations in some cases for improved results. 

Deep latent variable models, and by extension, variational autoencoders, can easily extend to the generation of different variable types by adjustment of the choice of distribution for each variable's likelihood term. This approach is an application of deep exponential families \citep{ranganath2015deep}.
Similarly, GAN-based approaches can handle different variable types, and several variants to improve performance in the tabular data setting have been proposed. 

{\em MEDGAN} \citep{choi2018generating} involves an autoencoder as part of the overall generative model, designed to improve performance on categorical and continuous data by first learning a continuous embedding of the discrete data. Several different refinements of a basic GAN approach to tabular data are considered in the \emph{conditional tabular GAN} (CTGAN) approach \citep{xu2019modeling}, namely {\em mode-specific normalization} and {\em conditioning vectors}. The latter approaches are applied to VAE models in the same work, yielding the Tabular VAE (TVAE) approach. The mode-specific normalization aims to address deficiencies of the usual standardization (normalization) preprocessing step in the case where univariate multimodal distributions exist for the individual variables in the dataset. The approach involves fitting a separate Gaussian mixture model to individual columns corresponding to continuous variables and subsequently using the parameters of a randomly-drawn mixture component to determine the parameters of a standardization transform applied as a preprocessing step during training. \cite{li2021inverse} propose {\em inverse-CDF GAN}, which replaces the mode-specific normalization in CTGAN with a transformation based on feeding each column of continuous data through an estimated cumulative distribution function, thus approximating the probability integral transform, which would yield samples uniformly on the interval between zero and one. A shift and scale transform is applied to make each column take values in the interval $(-1,1)$. The Inverse-CDF GAN approach yields improvements over CTGAN and TVAE in some experiments.

In GAN-based approaches, discrete variables such as count data are often treated as categorical  (e.g., variables corresponding to count data may be assumed only to take values matching the observed data) or continuous. However, the data set typically has a considerable class imbalance in the former case. Training models capable of generating synthetic data with categorical variables with class imbalances or many classes is challenging. CTGAN uses a {\em conditional generation} approach to address this challenge which involves assigning each generator sample a \emph{conditioning vector} during training. The conditioning vector is drawn by randomly sampling a categorical variable from all possible categorical variables in the data set and then sampling a value for the chosen variable according to its associated empirical probabilities. A modified loss function is used that encourages the conditioning vector and generated samples to match; see \cite[Section 4.3]{xu2019modeling} for details.

Adapting the loss function(s) to improve performance is a commonly employed technique, as often this can input some inductive bias into the model training procedure. Inverse-CDF GAN features a {\em label reconstruction function} as an additional output of its discriminator. 
For each batch of data given to the discriminator during training, a randomly-chosen variable is designated as the \emph{label} and a regression model incorporated into the discriminator predicts the label given the other variables. The discriminator's loss function includes an additional term that accounts for the accuracy of the predictions from the label reconstruction function. This approach provides the GAN with an additional way to learn the dependency structure in the dataset. TableGAN \citep{park2018data} proposes two additional loss functions, the \emph{information loss} and the \emph{classification loss}. The information loss compares the mean and variance of the marginal densities of synthetic samples with the observed data samples. In contrast, the classification loss is equivalent to the label reconstruction function described above.

There are two notable extensions of VAEs for the tabular data setting. The first is \textit{Oblivious VAEs} \citep{vardhan2020synthetic}, which incorporates differentiable oblivious decision trees (DODTs) \citep{popov2019neural} --- a neural network architecture explicitly designed for tabular-like datasets --- within VAEs. Oblivious decision trees (ODTs) \citep{langley1994oblivious}) are levelled graphs that constrain all nodes at a level to be split by the same variable, defining an ordering of the variables. Inference for an ODT is efficient because the computation of the splits occurs in parallel instead of sequentially, as in non-oblivious decision trees \citep{quinlan1986induction}. DODTs are a differentiable extension to ODTs, thus amenable to training through gradient-based optimization algorithms. The Oblivious VAE method outperformed the TVAE approach in some examples. The most recent refinement of VAE-based approaches to synthetic data based on relational modelling and graph neural networks (e.g., \cite{zhou2020graph}) appears in the GOGGLE model of \cite{liu2022goggle}.

Finally, it is worth noting that compared to GANs and VAEs, there is limited work regarding other model types investigating the challenges of tabular data. However, for normalizing flows, \cite{amiri2022generating} explore the efficacy of flow-based models with different base distributions in modelling heavy-tailed data for synthetic data generation, finding that a Student-$t$ or a Gaussian Mixture model as a base distribution tend to outperform the default choice of a standard normal in their experiments.

\section{Evaluation of Synthetic Datasets and DGMs}\label{synthetic_data_evaluation}
Evaluation methods for generative models often involve statistical comparisons between synthetic and observed datasets. One such method includes calculating the similarity score, as proposed by \citet{brenninkmeijer2019generation}. The score is derived from the Spearman correlation of specific quantities, such as means or standard deviations, from the marginal distributions of both datasets. In the context of tabular data, a commonly observed evaluation method is the {\em machine learning efficacy} approach, as described by \cite{xu2019modeling}. This method involves training two distinct predictive models, one on samples from the generative model and the other on the original data. A chosen variable within the dataset serves as the target response variable for these models. The models' predictive performances are compared using a test set from the observed dataset. However, it is crucial to note that the machine learning efficacy approach primarily assesses the ability of a classifier to explain variability in a particular {\em conditional} distribution related to the dataset. This approach may not comprehensively evaluate the alignment between the joint distribution of the generative model and the empirical data distribution. Therefore, while it can be useful, it may only partially capture the extent to which a synthetic data set can be used as an effective proxy for the real data set in all downstream tasks. 

The evaluation of probabilistic models, in standard settings, often involves a comparison between models via quantities related to the (marginal) likelihood $p_{\v \theta}(\v x)$ over all data under the fit model.
However, such quantities are often not directly available and must be estimated. For deep latent variable models, unbiased marginal likelihood estimation is a complex problem. However, it can be attempted via advanced Monte Carlo techniques, such as Annealed Importance Sampling \citep{neal2001annealed}.

However, direct unbiased estimation of $p_{\v \theta}$ is impossible for models such as GANs. It is possible, however, in such instances to use parzen windows \citep{parzen1962estimation} to approximate the likelihood (and by extension, the log-likelihood) for a set of samples using a kernel density estimate, typically with a Gaussian kernel. These windows can be fitted to both the observed dataset and the samples from the generative models of interest, allowing for a comparison of the average log-likelihoods of the respective sample sets. While these estimates only provide approximations, they potentially are useful if the ordering of average log-likelihoods given by this approximation matches the true ordering of log-likelihoods across models. \citet{bachman2015variational} discuss that the latter condition is not always satisfied. In some cases, the average log-likelihood estimated by a Parzen window for samples from a fitted model returns higher values than the average log-likelihood estimated for samples from the true data-generating process. Furthermore, comparing average log-likelihood values over generated data sets does not explain how or where the generative model's samples differ from the observed data and therefore is not particularly instructive in uncovering deficiencies. 
It is also worth noting that using the likelihood of generative models for evaluation may lead to misleading results in high-dimensional settings. \cite{theis2015note} discuss an illustrative example involving a two-component mixture comprised of the true data-generating distribution and a multivariate standard Gaussian (i.e., noise) distribution. The log-density of the true data distribution scales with dimension, while the random noise does not. Additionally, choosing different mixture weights makes a negligible difference in log-likelihood, even when the dimensionality of the generative model scales. An additional weakness of likelihood-based evaluation approaches is that they provide no guidance on \emph{how} data from a generative model differs from the true data distribution.

The above issues have inspired the development of alternatives to likelihood-based approaches.
\citet{sajjadi2018assessing}
propose \em{precision} and \em{recall} metrics for distributions. \cite{kynkaanniemi2019improved} propose an alternate approach to calculating precision and recall metrics that involve nonparametrically modelling the manifolds that the synthetic and observed samples approximately lie on. \citet{naeem2020reliable} argue that such metrics do not satisfy certain desirable properties that we would like to see in evaluation metrics for generative models and propose the metrics {\em density} and {\em coverage}, which are argued to be more reliable alternatives. \citet{alaa2022faithful} propose $\alpha$\emph{-precision}, $\beta$-{\em recall}, and the {\em authenticity} metric. The latter aims to evaluate whether generated samples appear to memorize or are similar to the observed (training) data points. As $\alpha$-precision, $\beta$-recall, and authenticity are all defined at the individual-sample level, they can be used to remove particular generated samples. Experiments performed by \cite{alaa2022faithful} show that such post hoc corrections improve the performance of any generative model. 

{\em Siamese neural networks} \citep{lecun2005loss, chopra2005learning} are pairs of neural networks that share parameters trained via an objective involving something called a \emph{contrastive loss} \citep{wang2020understanding} to return outputs, called {\em feature vectors}, that are similar if the inputs to the two neural networks are considered similar, and differ considerably otherwise. Siamese neural networks can be used to train a classification model on the observed data, which can be used to evaluate generative models using the Siamese distance score (SDS, \cite{torfi2021evaluation}) under the assumption that samples from the observed dataset have some notion of a label or grouping. SDS computes an estimate for the label of each synthetic sample, using a majority vote of the labels of some number of the closest observed samples in the feature space produced by the Siamese neural network. For each synthetic sample, we can then compute the average distance given by the output of the Siamese neural network between the sample and the observed samples that share the same label. Averaging over this value for all synthetic samples gives the Siamese distance score.

\section{Private Machine Learning and Differential Privacy}\label{differential_privacy_ml}
Generating synthetic data through deep generative models is a promising solution aimed at safeguarding the privacy of the original data while maintaining its inherent features. However, these approaches often fail to provide explicit privacy assurances, indicating a need for methods to ensure data confidentiality during synthetic data generation. This need has spurred the growth of ``private machine learning'', an emerging field dedicated to developing machine learning and probabilistic models that can handle confidential datasets while offering quantifiable privacy guarantees. The bedrock of this field lies in a mathematical framework for data privacy, known as \emph{differential privacy} (DP) \citep{dwork2014algorithmic}.

This section provides an overview of DP, including its various forms, properties, and applications in deep generative models. We aim to explain the fundamental aspects of DP and its application in recent differentially private model training methods. For a more in-depth discussion, readers can refer to the monograph by \cite{dwork2014algorithmic} and the recent survey by \cite{wang2023differential} that centres on deep learning. Simple software demonstrating the effect of different degrees of enforced differential privacy is available, courtesy of \citet{aitsam2022differential}.

\subsection{Differential Privacy: An Overview}\label{differential_privacy}
Before we delve into the different forms of DP, it is worth noting that the formal definition of DP and its underlying concept relates to a specific \emph{type} of privacy that has considerable advantages in providing a formal framework for the modular design of privacy-preserving algorithms that admit tractable mathematical analysis, as will be demonstrated shortly.

Defining the setting and key terms is essential to discuss the different types of DP. Consider a dataset $\m X \in \bb R^{N\times P}$. We want to release $f({\rm X})$. Here, $f$ can be any function of the dataset, for example, the dataset itself, a summary statistic or vector of summary statistics, or model parameters corresponding to some model trained using $\m X$. Differential privacy involves releasing such information in a way that involves randomness --- by applying a ``random mechanism'' $M$ to the data, where the mechanism $M$ implicitly involves the function $f$. The properties of this randomness --- the output of the random mechanism, $M(\m X)$ --- form the crux of differential privacy.

\subsubsection{Pure Differential Privacy}\label{pure_differential_privacy} Let $\mathcal{S}$ denote the image of $M$, i.e., the set of all possible outcomes obtained by applying $M$. For tabular data, we designate a dataset $\mathrm{X}'\in \mathbb{R}^{N\times P}$ as a {\em neighbour} of $\mathrm{X}$ if it is a dataset $\mathrm{X}'$ of equivalent dimensions to $\mathrm{X}$, differing from $\mathrm{X}$ in merely {\em one} row. This condition involves removing and substituting data corresponding to a {\em single} observation.

A randomized algorithm $M$ satisfies $\epsilon$-\emph{differential privacy} ($\epsilon$-DP) for a given $\epsilon>0$ if, for all $S\subseteq \mathcal{S}$ and any pair of neighboring datasets $\m X$ and $\m X'$, it holds that
\[\bb P(M(\mathrm{X}) \in S) \le \exp( \epsilon){\bb P(M({\rm X}')\in S)}.\] 
This condition states that the probabilities of potential outcomes under $M(\m X)$ do not vary substantially when any observation in $\m X$ is substituted. Smaller values of $\epsilon$ denote higher privacy levels.

To facilitate understanding, consider the following. As $\epsilon \to 0$, a mechanism $M$ adhering to $\epsilon$-DP implies that $\bb P(M(\mathrm{X}) \in S) \le \bb P(M({\rm X}')\in S)$, which further suggests $\bb P(M(\mathrm{X}) \in S) = \bb P(M({\rm X}')\in S)$. Hence, the probability of obtaining certain outcomes from $M(\m X)$ does {\em not} significantly change if any single observation (row) of $\m X$ is replaced.

\subsubsection{Approximate Differential Privacy}
\emph{Approximate differential privacy}, $(\epsilon, \delta)$-DP, is a relaxation of pure differential privacy, which involves an additional so-called \emph{slack parameter} $\delta \in (0,1)$. A randomized algorithm $M$ is said to be $(\epsilon, \delta)$-DP, if for any $S\subseteq \mathcal{S}$ and any two neighboring datasets $\m X$ and $\m X'$, it holds that \[\bb P(M(\mathrm{X}) \in S) \le \exp( \epsilon){\bb P(M({\rm X}')\in S)} + \delta.\] 

The introduction of the slack parameter allows 
for the existence of sets $S^* \subset \mathcal{S}$ for which $\bb P(M(\m X) \in S^\star) \le \delta$ but $\bb P(M(\m X') \in S^\star) = 0$ for at least one $\m X'$ neighbouring $X$. For such sets, the random mechanism essentially fails, so the slack parameter is often referred to heuristically as the ``probability of \emph{privacy failure} of $M$''.

The advantage of allowing a slack parameter is that algorithms that satisfy approximate differential privacy require less injected noise into the mechanism than those that satisfy pure differential privacy for the same $\epsilon$ (e.g., by only requiring the addition of noise from the Gaussian rather than the Laplace distribution).

\subsubsection{Additive Noise Mechanisms and Sensitivity}
A commonly-used class of mechanism for $f$ that map input data sets to $\bb R^d$ for some $d\in \bb N$ is that of {\em additive noise} (or {\em noise-adding}) mechanisms. Here, $M(\m X) = f(\m X) + \v z$,
where $\v z$ is a vector of independent and identically-distributed draws from some noise distribution. The {\em Laplace Mechanism} takes the elements of $\v z$ as distributed according to a zero-mean Laplace distribution with some specified scale parameter. The {\em Gaussian Mechanism} instead has elements of $\v z$ that are Gaussian distributed. 

Note that any mechanism's privacy properties rely on $f$. A commonly-used quantity in the literature is the {\em global  sensitivity} of $f$, $\Delta_p^{(f)}$ denoted which is defined as 
$\Delta_p^{(f)} = \max_{X, \m X'}||f({\rm{X}})-f(\mathrm{X}')||_p$, where $||\cdot ||_p$ denotes the $p$-norm for some $p \ge 1$, and the maximum is defined over all pairs of neighbouring data sets. For arbitrary $\epsilon \in (0,1)$, it can be shown \cite[Theorem A.1]{dwork2014algorithmic} that, for $c^2 > 2\log(1.25/\delta)$, the Gaussian mechanism with scale parameter $\sigma \ge c \Delta_2^{(f)} \epsilon^{-1}$ is $(\epsilon,\delta)$-DP.

Note that the sensitivity may be unbounded; for example, if $f(\m X)$ returns the mean for one column of $X$, then the sensitivity is infinite (as one row can plausibly take any value in the corresponding column). For reasons such as this, it is common in the literature to involve \emph{clipping} as part of $f$ to ensure the sensitivity is finite and can be calculated to derive privacy bounds. 

\subsubsection{Renyi Differential Privacy}\label{renyi_differential_privacy}
\emph{R\'enyi Differential privacy} (RDP,  \cite{mironov2017renyi}) is an alternate privacy measure. Like $(\epsilon, \delta)$-DP, RDP relaxes the constraints of $\epsilon$-DP but offers certain advantages over $(\epsilon, \delta)$-DP for theoretical reasons surrounding its associated mathematical analysis, discussed shortly. 
The R\'enyi divergence is so-named because it is inspired by the classical Renyi divergence \emph{R\'enyi divergence} \citep{renyi1961measures} between two probability distributions $p$ and $q$: 
$\bb D_{\alpha}(p||q) = \frac{1}{\alpha-1}\log \mathbb{E}_{q}\big(p(\v x)/{q(\v x)}\big)^\alpha$. As $\alpha\rightarrow 1$, the Renyi divergence converges to the Kullback--Leibler divergence $\bb K \bb L(p||q)$. 

Write $M_X$ to denote the distribution of the object obtained by applying $M$ to some base input $\m X$. For $\alpha > 1$, a randomized algorithm $M$ satisfies $(\alpha, \epsilon)$-RDP, if for all possible $S \subseteq \mathcal{S}$ generated by applying $M$ to any pair of neighbouring data sets $\m X, \m X'$, it holds that $\bb D_{\alpha}(M_\m X || M_{\m X'})\leq \epsilon$. 
Renyi differential privacy is related to both pure and approximate differential privacy. As $\alpha\rightarrow \infty$, $(\alpha, \epsilon)$-RDP becomes equivalent to $\epsilon$-DP. Moreover, any mechanism that is $(\alpha, \epsilon)$-RDP also satisfies $\left(\epsilon + {\log ({1}/{\delta})/{(\alpha-1)}}, \delta\right)$-DP for {\em any}  $\delta\in (0, 1)$ \cite[Proposition 3]{mironov2017renyi}.

\subsubsection{Properties of differential privacy}\label{dp_properties}
Differential privacy has several important and beneficial properties, which we now discuss.

\emph{Group privacy} extends privacy to a subset of $k$ observations. If a mechanism satisfies $\epsilon$-DP (or $(\epsilon, \delta)$-DP), the privacy bound for any set of $k$ observations satisfies $k\epsilon$-DP (or $(k\epsilon, \delta)$-DP). Differential privacy also \emph{future proofs} the original dataset, meaning that after a differentially private data release, post-processing of that data or additional data releases from auxiliary datasets will not impact the differentially private nature of the initial release.

However, multiple private data releases using the same dataset alter the privacy bound. An essential property of differential privacy is its \emph{composition}, enabling the calculation of an updated privacy bound when combining multiple differentially private mechanisms. A basic composition result is that if mechanisms $M_1$ and $M_2$ satisfy $\epsilon_1$-DP and $\epsilon_2$-DP, respectively, then $M = (M_1,  M_2)$, where the input of $M_2$ may optionally also depend on the output of $M_1$,  is $(\epsilon_1 + \epsilon_2)$-DP. The composition properties of approximate differential privacy are more involved than the composition properties of pure differential privacy. A composition of $k$ $(\epsilon, \delta)$-DP mechanisms satisfies $(k\epsilon, k\delta)$-DP \citep{dwork2009differential, dwork2010boosting}, but this is not a tight bound. An optimal privacy bound for composing $k$ $(\epsilon, \delta)$-DP mechanisms exists \citep{kairouz2017composition}, but is challenging to compute in practice. A significant advantage for R\'enyi differential privacy is that it admits results allowing tighter analysis for bounds relating the composition of $k$-RDP mechanisms in several settings \citep{mironov2017renyi}.

A final important property is \emph{post-processing}: the mechanism obtained by applying an arbitrary function $g$ to the output of a differentially-private mechanism $M$ satisfies the same DP properties as $M$. This property is significant for parameter estimation in machine learning and probabilistic models. Post-processing plays a crucial role in training private machine learning models. The post-processing property ensures simulated output (i.e., synthetic data) from the model trained to have parameters obtained from a differentially-private mechanism satisfies DP.

\subsection{Private Machine Learning}
This section explores methods that leverage fundamental properties of differential privacy for training machine learning models. The two primary approaches to ensure differentially private output from a model are {\em private training} \citep{abadi2016deep, mironov2017renyi, papernot2018scalable} and {\em private prediction} \citep{dwork2018prediction, bassily2018agnostic}. Private training guarantees that the model's parameters satisfy differential privacy, while private prediction enables users to interact with trained models by submitting inputs, ensuring that instead, the model's output is differentially private. As the latter is not relevant to the task of synthetic data generation, this section focuses exclusively on the task of private training. 

\subsubsection{Differentially Private Stochastic Gradient Descent}\label{section:dp_sgd}
Differentially private stochastic gradient descent (DP-SGD) preserves differential privacy within a stochastic gradient descent (SGD) algorithm \citep{song2013stochastic, bassily2014private}. In a single iteration of DP-SGD, a random subsample of observations is chosen, and the gradients corresponding to the model parameters are calculated for each observation. The gradient vectors are then \textit{clipped} elementwise. Following this, the individual clipped gradients are averaged, and Gaussian noise is added. This gradient step costs $(\mathcal{O}(q\epsilon), q\delta)$-DP through a result known as the \textit{privacy amplification lemma} \citep{kasiviswanathan2011what, balle2018privacy}, where $q$ is the probability of selecting each observation as part of the subsample for any given iteration. Intuitively, as the subsample size remains constant and the number of data observations increases, the gradient update's confidentiality should be considered greater, resulting in a lower privacy cost. For these algorithms to be practically helpful, theoretical results are needed to determine how the privacy cost is composed as the subsampled Gaussian mechanism is iteratively applied within a gradient descent-style algorithm. \cite{abadi2016deep} derive the properties of Poisson subsampling under a sequence of Gaussian mechanisms, while \cite{wang2019subsampled} derived the properties of subsampling without replacement under a sequence of Gaussian mechanisms. A unified perspective on privacy amplification techniques for different subsampling and neighbouring relations can be found in  \cite{balle2018privacy}. 

Using the composition property of DP for a sequence of $T$ steps in a gradient descent algorithm establishes that DP-SGD is $\left(\mathcal{O}(qT\epsilon), qT\delta\right)$-DP. Recall that such results are based on bounds that are not tight. To overcome this issue, a result called the \textit{strong composition theorem} \citep{dwork2010boosting} allows for tracking a sequence of subsampling Gaussian mechanisms and provides a tighter bound of $\left(\mathcal{O}(q\sqrt{T}\epsilon), \mathcal{O}(q T\delta)\right)$-DP for $T$ iterations of an $\left(\mathcal{O}(p\epsilon), p\delta\right)$-DP algorithm.

The {\em moment's accountant} \citep{abadi2016deep} uses algorithm-specific properties to track the privacy loss of compositions of subsampled Gaussian mechanisms. By considering a random variable called the {\em privacy loss}, and monitoring its logarithmic moments, DP-SGD satisfies the privacy bound $(\mathcal{O}(q\epsilon\sqrt{T}), \delta)$-DP. Implementing the moment's accountant focuses on the log of the moment generating function (MGF) of the so-called privacy loss random variable arising from the mechanism. For a given mechanism $M$, we denote this moment generating function, which is computed via numerical integration in practice, by $\alpha_M(\cdot)$. Two properties of $\alpha_M$ are established \cite[Theorem 2]{abadi2016deep}. Firstly, if $M$ is a sequence of randomized mechanism $M_1, \ldots, M_k$, the bound on the log-MGF is additive, $\alpha_M(\lambda)\leq\sum_{i=1}^k \alpha_{M_i}(\lambda)$. Secondly, for arbitrary $\epsilon > 0$, the randomized algorithm $M$ satisfies $(\epsilon, \delta)$-DP for $\delta = \min_\lambda \exp(\alpha_M(\lambda)-\lambda\epsilon)$. At the $t$-th step, the DP-SGD algorithm calculates the bound for $\alpha_{M_t}(\lambda)$ at fixed values of $\lambda$. The sum of these bounds creates a bound for $\alpha_M(\lambda)$ for each value of $\lambda$. The second property above then allows for converting the bounds for the moment into an $(\epsilon, \delta)$-DP privacy guarantee. Given a fixed value of $\delta$, one can search through the values of $\lambda$ to find the minimum value of $\epsilon$, or vice versa.

The \textit{analytical moment's accountant} \citep{wang2019subsampled} tracks the privacy loss of the composition of subsampling mechanisms using R\'enyi differential privacy. The analytical moment's accountant can monitor the privacy loss involving any mechanism that satisfies RDP. In contrast, the original moment's accountant can only track the bound on sub-sampled Gaussian mechanisms. As the analytical moment's accountant can store the log of the MGF in symbolic form under the definition of RDP, it can track the log of the MGF for all values $\lambda \geq 1$ and does not require numerical integration. By leveraging the relationship between $(\alpha, \epsilon)$-RDP and $(\epsilon, \delta)$-DP, the privacy loss tracked under RDP can be easily converted to $(\epsilon, \delta)$-DP for any choice of $\delta$ or $\epsilon$.

\subsubsection{Private Aggregation of Teacher Ensembles}

{\em Private Aggregation of Teacher Ensembles} (PATE) \citep{papernot2017semisupervised, papernot2018scalable} offers an alternative to DP-SGD-style algorithms for estimating private parameters of machine learning models that involve a classifier as part of the model (e.g., GANs). For the latter reason, the approach is thus considerably less generally-applicable than DP-SGD, but still worth mentioning.

The PATE approach involves three aspects: an ensemble of \textit{teacher models}, an \textit{aggregation mechanism}, and a \textit{student model}. The approach involves partitioning a dataset into $K$ disjoint subsets (i.e., each observation belongs to one subset), and then fitting a teacher model for each data partition independently, resulting in $K$ teacher models. All teachers are the same model type, and there are no restrictions on the algorithms used to fit the teacher models (i.e., differentially-private training methods do not need to be used). As the general approach is based on a classifier model, we describe it in that context here, but extensions to generative models are discussed in the following subsection. 

Denoting the count of teacher predictions for each class $j$ after taking an ensemble of teacher predictions for input $\v x$ by $n_j(\v x)$. \citet{papernot2017semisupervised} employ the \textit{noisy-max aggregation mechanism} $M_b(\v x)$ of the teacher predictions, \[M(\v x) = \text{arg }\max_j\left\{n_j(\v x)+\text{Laplace}\left({1/\gamma}\right)\right\},\] with privacy parameter $\gamma$ and $\text{Laplace}(b)$ representing noise from the Laplace distribution with location $0$ and scale $b$. \cite{papernot2018scalable} extends this aggregation method by proposing the \textit{Gaussian noisy max} (GNMax) aggregation mechanism and study its properties. GNMax replaces the Laplace distribution in the noisy-max aggregation mechanism with a Gaussian distribution.

The final aspect of the PATE approach is the use of a {\em student model}, designed to overcome two privacy concerns with the above aggregation mechanisms. Firstly, every additional prediction released decreases the overall privacy level. Secondly, whilst the individual predictions may be differentially-private, the original teachers' models on which they are based are not, precluding their release if privacy guarantees are required. To address the issues mentioned above, the final element of the PATE approach involves training the student model, chosen to be a semi-supervised variant of the Generative Adversarial Network (GAN, Section \ref{gan}) shown to improve semi-supervised learning by \cite{salimans2016improved}, using unlabeled data that is non-sensitive, some of which is labelled using the aggregation mechanism from the teacher model ensemble. Consequentially, when deploying the student model, only the privacy properties arising from training the student model need to be considered, which do not increase with the number of predictions. Privacy is preserved because the student model's training does not use private data, even if the architecture and parameters of the student model are public. Two data-dependent adaptations of the GNMax aggregation mechanism are \textit{confident-GNMax aggregation} and \textit{interactive-GNMax aggregation} \citep{papernot2018scalable}.

\subsection{Private Deep Generative Models}\label{dp_machine_learning_models}
The adoption of DP-SGD and related methods for privately training deep generative models has been greatly facilitated by a combination of its general applicability and the availability open-source differentially private optimization libraries such as {\textsf{Opacus}} \citep{yousefpour2021opacus} for {\sf PyTorch} \citep{paszke2019pytorch}, {\sf Optax} \citep{bradbury2018jax} for {\sf JAX} \citep{deepmind2020jax}, and native support in {\sf TensorFlow} \citep{abadi2016deep}. In the literature, differential privacy-enabled training has been considered for many of the models that were discussed in Section \ref{generative_models}, including normalizing flows \citep{waites2021differentially, lee2022differentially}, variational autoencoders \citep{chen2018differentially}, and GANs. There has been a particular focus on the latter. Two notable applications of differentially-private GAN approaches for health data include the generation of heterogeneous electronic health records \citep{chincheong2019generation} and synthetic patient-level systolic blood pressure samples \citep{beaulieu2019privacy}.

For a comprehensive review of the differentially-private GAN approaches and aspects surrounding their training, we refer to \cite{fan2020survey}. While DP-SGD is often easily applied, a central idea in the literature for constructing differentially-private training schemes for DGMs is exploring how to ensure a high level of model performance for a given privacy budget. \cite{zhang2018differentially} consider a ``warm-starting'' approach where an initial phase employs public data, after which differentially-private updates are used with sensitive data, the key idea being that fewer steps of DP-SGD as a result of the warm start will yield tighter privacy bounds. 
One interesting idea is that not all parameters necessarily need to be updated by DP-SGD. Examples include latent variable models (Section \ref{latent_variable_models}) and generative adversarial nets (Section \ref{gan}), where the training procedure involves iteratively updating two distinct sets of parameters by gradient descent until the objective function converges. One set, $\v\theta$, parameterizes the generative model. In contrast, the other parameters, consisting of the variational parameters $\v\phi$ for latent variable models or the discriminator parameters $\v \lambda$ for GANs, do not appear in the generative model. Some methods propose that if not all parameters will be publicly released, it is only necessary to update the generative model parameters by DP-SGD. This idea appears in \emph{gradient-santitized Wasserstein GAN} (GS-WAN, \cite{chen2020gswgan}) and {\em differentially-private tabular GAN} (DTGAN, \cite{kunar2021dtgan}). An opposite approach relies on observing that updates of the generative model parameters $\boldsymbol{\theta}$ depend on parameters $\v \lambda$, and thus it may be prudent to update $\v \lambda$ using DP-SGD. The intuition is that updates of the generative model parameters satisfy differential privacy via the post-processing property because they depend on differentially private values of $\v \lambda$. \citet{kunar2021dtgan} explore and make a case for the latter strategy.
 As discussed in Section \ref{tabular_generative_modeling}, some  generative models involve learning an {\em encoding} of the data, an idea that has been explored in the differentially-private setting by {\em differentially-private conditional GAN} \citep{Tanti2021} and {\em differentially-private convolutional GAN} \citep{torfi2022differentially}

Whilst DP-SGD is the typical approach, some methods train differentially private GANs via the PATE approach. PATEGAN \citep{jordon2018pate} involves training a student discriminator by maximizing the binary cross-entropy of the information given by multiple teacher models. The training of PATEGAN contrasts with the semi-supervised approach using publicly available data to train the student model in the original PATE implementation. \citet{long2021gpate} notes that only the generator must satisfy privacy, allowing the student discriminator to observe real samples during training, yielding the G-PATE approach. Additionally, \citet{long2021gpate} proposes a private aggregation mechanism applicable to any variable type, not just categorical variables.

\section{Discussion}\label{discussion}
The paper provided a cohesive overview of deep generative models for synthetic data generation, focusing on the under-explored domain of tabular data. We have discussed the flexibility and potential uses of neural networks in probabilistic models, the various deep generative models employed for synthetic data generation, their inference algorithms, their adaption to tabular datasets, considerations in privacy-sensitive settings, and the evaluation metrics for assessing the fidelity of high-dimensional synthetic datasets.

While deep generative models have gained immense popularity and provided state-of-the-art results for image and text datasets, they are often criticized for their poor performance and excessive complexity in less common machine learning use cases, such as medium-sized or smaller tabular datasets. However, we argue that no inherent limitation in these models would hinder their ability to excel in such scenarios. Overfitting and the lack of generalization are common concerns in deep learning methods due to the heavily parameterized nature of neural networks. However, in synthetic data generation, these concerns could be more relevant as our goal is to estimate and sample from the data-generating process of the observed dataset. It is \textit{precisely}, because we aim to capture the data-generating process, that privacy-aware parameter estimation becomes crucial for sensitive applications.

We believe that the potential of deep generative models in generating and releasing synthetic versions of confidential data sources has only begun to be explored, and note that recently-available software package {\sf Synthcity} \citep{qian2023synthcity} serves as a promising enabler for experimentation for several approaches discussed herein. 

There are numerous exciting avenues for methodological development, such as scalable nonlinear extensions to existing generative modelling approaches, private inference for probabilistic models like unsupervised clustering techniques, and scalable approaches to model fitting and comparisons. These advancements will pave the way for future applications involving confidential data sources. The article focused on the most prominent generative models. It is worth noting, however, that in recent years a class of generative models called {\em diffusion models} has enjoyed considerable success in domains outside of the tabular data setting; for two survey articles, see \citet{yang2022diffusion} and \citet{croitoru2023diffusion}. Their application to tabular data and appropriate modifications required for such settings are interesting potential avenues of research worthy of exploration.

We highlight that the marriage of deep learning and traditional statistical approaches opens up cross-disciplinary collaboration and innovative opportunities. By combining the strengths of both fields, we can devise novel techniques that address the limitations of current methods and provide more robust, accurate, and privacy-sensitive synthetic data generation. As the demand for data-driven solutions grows across various industries, privacy-preserving synthetic data generation will become more critical. By harnessing the power of deep generative models, we can unlock the full potential of confidential data sources, thereby enabling the broader exploration, application building, and methodological development that will ultimately contribute to advancing machine learning, artificial intelligence, and data-driven decision-making.

\bibliographystyle{plainnat}
\bibliography{ref.bib}
\end{document}